\documentclass[letterpaper, 10 pt, conference]{ieeeconf} 

\IEEEoverridecommandlockouts 
\usepackage{graphics} 
\usepackage{subcaption}
\usepackage{amsmath} 
\usepackage{amssymb} 
\usepackage{mathtools}
\usepackage{bm}
\usepackage[ruled, vlined]{algorithm2e}
\usepackage{booktabs}
\usepackage{caption}
\usepackage{multirow}
\usepackage{siunitx}

\title{\LARGE \bf GraSP-STL: A Graph-Based Framework for Zero-Shot Signal Temporal Logic Planning via Offline Goal-Conditioned Reinforcement Learning}

\author{Ancheng Hou, Ruijia Liu and Xiang Yin
\thanks{This work was supported by the National Science and Technology Major Project (2025ZD1600700) and the National Natural Science Foundation of China (62573291,62533017,62173226).}
\thanks{A. Hou, R. Liu and X. Yin are with the School of Automation \& Intelligent Sensing, Shanghai Jiao Tong University, Shanghai 200240, China. {\tt\small \{hou.ancheng, liuruijia, yinxiang\}@sjtu.edu.cn}}%
}

\begin{document}
   \maketitle
   \thispagestyle{empty}
   \pagestyle{empty}

   \begin{abstract}
      This paper studies offline, zero-shot planning under Signal Temporal Logic (STL) specifications. We assume access only to an offline dataset of state-action-state transitions collected by a task-agnostic behavior policy, with no analytical dynamics model, no further environment interaction, and no task-specific retraining. The objective is to synthesize a control strategy whose resulting trajectory satisfies an arbitrary unseen STL specification. To this end, we propose GraSP-STL, a graph-search-based framework for zero-shot STL planning from offline trajectories. The method learns a goal-conditioned value function from offline data and uses it to induce a finite-horizon reachability metric over the state space. Based on this metric, it constructs a directed graph abstraction whose nodes represent representative states and whose edges encode feasible short-horizon transitions. Planning is then formulated as a graph search over waypoint sequences, evaluated using arithmetic-geometric mean robustness and its interval semantics, and executed by a learned goal-conditioned policy. The proposed framework separates reusable reachability learning from task-conditioned planning, enabling zero-shot generalization to unseen STL tasks and long-horizon planning through the composition of short-horizon behaviors from offline data. Experimental results demonstrate its effectiveness on a range of offline STL planning tasks.
   \end{abstract}

   \section{Introduction}
   Signal Temporal Logic (STL) is a formal language for specifying complex temporal tasks for dynamical systems with both temporal and spatial constraints, allowing the formalization of objectives such as sequencing, timing, and safety requirements. By leveraging formal reasoning, STL enables the synthesis of control inputs or plans that guarantee the satisfaction of these specifications with provable correctness. In recent years, STL has been widely applied across various engineering domains where precise specification and reliable execution of temporal behaviors are critical~\cite{yin2024formal,gu2025robust}.

   When an accurate system model is available, optimization-based methods are a standard approach for STL control synthesis~\cite{raman2014model,belta2019formal}. By encoding specifications as mixed-integer or nonlinear programs, they systematically handle logical and temporal constraints. Despite numerous improved variants~\cite{kurtz2022mixed,sun2022multi}, these methods often face scalability issues as the state dimension, planning horizon, or specification complexity increases. Sampling-based methods, such as RRT~\cite{vasile2017sampling,ahmad2026rrt}, partially address this by exploring trajectories and evaluating STL robustness without explicit state-space discretization, but they still typically rely on analytical dynamics models or forward simulations, which may be unavailable or difficult to construct.

   Learning-based methods offer a promising alternative for STL control synthesis without explicit system models. A common approach uses reinforcement learning (RL), encoding STL objectives into reward functions and optimizing policies through interaction with the environment~\cite{venkataraman2020tractable,saxena2023funnel,wang2025multi}. While effective in some cases, RL depends on online data, making it unsuitable for purely offline settings, and learned policies are typically task-specific, limiting generalization. Long-horizon STL tasks remain challenging due to sparse and delayed rewards. End-to-end learning pipelines~\cite{hashimoto2022stl2vec,meng2023signal} can address offline scenarios by mapping observations directly to behaviors, but they often require substantial expert data and generalize poorly to unseen specifications.

   Recently, hierarchical planning frameworks~\cite{liu2025zero} have aimed to improve task generalization by integrating logic-level planning with learned models of system dynamics. By decomposing STL formulas into intermediate temporal subgoals and solving them with reusable learned modules, these methods achieve zero-shot generalization to unseen tasks without task-specific retraining. Nonetheless, they often entail high computational costs and require sufficiently rich data to support subgoal transitions, making them less practical when only fragmented offline trajectories are available.

   In this work, we focus on a model-free offline setting for STL planning, where no analytical system model or online environment interaction is available. Instead, we assume access only to a collection of offline system trajectories, potentially fragmented and collected via arbitrary task-agnostic policies. This setting presents two primary challenges. First, the planner must rely entirely on offline data, without querying the environment or refining decisions through online interaction. Second, long-horizon STL tasks are particularly difficult, as individual trajectory fragments are often short and may not directly satisfy the target specification. Addressing unseen STL tasks in a zero-shot manner thus requires inferring a reusable finite-horizon reachability structure from offline data to stitch short-horizon behaviors into coherent long-horizon plans.

   To address these challenges, we propose \textbf{GraSP-STL}, a graph-search-based framework for zero-shot STL planning from offline trajectories. It learns a goal-conditioned value function to induce a finite-horizon reachability measure and constructs a directed graph where nodes represent representative states and edges encode feasible short-horizon transitions. Planning a given STL specification is formulated as a graph-search problem over candidate waypoint sequences, with partial and complete plans evaluated using arithmetic-geometric mean (AGM) robustness~\cite{mehdipour2019arithmetic,ahmad2026rrt}. This enables long-horizon plans to be synthesized by stitching together short-horizon transitions from offline data, executed via a goal-conditioned policy. Our contributions are: (i) formalizing purely offline zero-shot STL planning with only fragmented task-agnostic trajectories, (ii) developing a graph-based reachability abstraction for long-horizon planning, (iii) integrating AGM robustness into graph search for structured STL planning, and (iv) demonstrating effective and interpretable solutions to unseen long-horizon STL tasks.

   \section{Preliminaries}

   \subsection{System Model}
   We consider a discrete-time control system with unknown dynamics described by the following form
   \begin{equation}
      \bm{x}_{t+1}= f\left(\bm{x}_{t}, \bm{a}_{t}\right),
   \end{equation}
   where $\bm{x}_{t}\in \mathbb{R}^{n}$ and $\bm{a}_{t}\in \mathbb{R}^{m}$ are the state and action at time $t$, respectively. Given an initial state $\bm{x}_{0}$ and a sequence of actions $\bm{a}_{0}\bm{a}_{1}\cdots \bm{a}_{T-1}$, the system generates a trajectory $\bm{\tau}= \left(\bm{x}_{0}\bm{x}_{1}\bm{x}_{2}\bm{x}_{3}\ldots \bm{x}_{T}\right )$ such that $\bm{x}_{t+1}=f(\bm{x}_{t}, \bm{a}_{t})$.

   To specify temporal logic constraints, we define a low-frequency signal by subsampling the system trajectory at a fixed rate. Let $k \in \mathbb{N}$ denote the number of control steps between two consecutive signal samples. Given a trajectory $\bm{\tau}= \left(\bm{x}_{0}\bm{x}_{1}\bm{x}_{2}\bm{x}_{3}\ldots \bm{x}_{T}\right )$, we define the associated signal $\bm{s}= \left(\bm{s}_{0}\bm{s}_{1}\bm{s}_{2}\bm{s}_{3}\ldots \bm{s}_{N}\right)$ as
   \begin{equation}
      \bm{s}_{i}\coloneqq
      \begin{cases}
         \bm{x}_{ik}, & \text{if }ik \le T, \\
         \bm{x}_{T},  & \text{otherwise},
      \end{cases}
      \quad i = 0, 1, \dotsc, N.
   \end{equation}
   where $N = \lceil T/k \rceil$. That is, the signal is obtained by sampling the system state every $k$ control steps, corresponding to a coarser task-level time scale, and the final state is repeated if necessary to cover the entire trajectory. This abstraction is common in practice, where high-frequency control is used for stabilization while task specifications are evaluated at a lower temporal resolution. We further denote partial signals by $\bm{s}_{t:t'}= \left(\bm{s}_{t}, \bm{s}_{t+1}, \ldots, \bm{s}_{t'}\right)$, where $0\le t \le t' \le N$.

   \subsection{Signal Temporal Logic}
   We adopt signal temporal logic (STL) specification \cite{stl} to describe formal temporal properties defined over the sampled real-valued signal $\bm{s}$. The syntax of STL formulae $\phi$ is:
   \begin{equation}
      \phi \coloneqq \top \mid \mu \mid \neg\phi \mid \phi_{1}\land \phi_{2}\mid \phi_{1}\lor \phi_{2}\mid \mathbf{G}_{[a,b]}\phi_{1}\mid \mathbf{F}_{[a,b]}\phi_{1},
   \end{equation}
   where $\top$ is the true predicate, $\mu$ is an atomic predicate associated with an evaluation function $h_{\mu}: \mathbb{R}^{n}\to \mathbb{R}$, i.e., predicate $\mu$ is satisfied at state $\bm{x}$ iff $h_{\mu}(\bm{x}) \ge 0$. The \emph{boolean operators} $\neg$, $\land$, and $\lor$ denote negation, conjunction, and disjunction, respectively. The \emph{temporal operators} $\mathbf{G}_{[a,b]}$ and $\mathbf{F}_{[a,b]}$ denote ``always" and ``eventually" over the time interval $[a,b]$, where $a,b \in \mathbb{Z}_{\ge 0}$ and $a \le b$.

   The boolean satisfaction of a given signal $\bm{s}$ at time $t$ with respect to an STL formula $\phi$ is denoted as $\bm{s}_{t}\models \phi$, and we further denote $\bm{s}\models \phi$ if $\bm{s}_{0}\models \phi$. Formal definitions of STL semantics can be found in~\cite{stl}.

   \subsection{AGM Robustness and Interval Semantics}
   In addition to boolean satisfaction, STL also admits a quantitative semantics known as robustness, which measures the degree to which a signal satisfies or violates an STL formula. The original robustness~\cite{robustness} of a signal $\bm{s}$ is defined over min-max operations, which can lead to non-smoothness and sensitivity to local extrema. In this work, we adopt the arithmetic-geometric mean (AGM) robustness semantics for STL from~\cite{mehdipour2019arithmetic}, which provides a smoother and more informative measure of satisfaction by replacing min-max with differentiable approximations. This property makes it particularly suitable for evaluating candidate trajectories during graph search.

   For a complete signal $\bm{s}$ and an STL formula $\phi$, let $\eta_{\bm{s},\phi}(t)\in[-1,1]$ denote the AGM robustness of $\phi$ at time $t$, where positive and negative values indicate satisfaction and violation, respectively. For predicate formulas, the evaluation function $h_{\mu}$ is normalized to $\widetilde{h}_{\mu}\in[-1,1]$, ensuring that the resulting robustness values remain within this range. When the signal is clear from context, we write $\eta_{\phi}(t)$ for brevity. When only a partial signal is available, we adopt the interval semantics in~\cite{ahmad2026rrt}. Specifically,
   \begin{equation}
      [\eta]_{\phi}(t)\coloneqq [\underline{\eta}_{\phi}(t), \overline{\eta}_{\phi}(t)] \subseteq [-1,1],
   \end{equation}
   denotes the range of AGM robustness values over all completions of the current partial signal. This interval semantics is sound, in the sense that it bounds the robustness of every completion, and it converges to the exact robustness once the observed signal length reaches the STL horizon $\lVert \phi \rVert$~\cite{ahmad2026rrt}.

   For temporal operators, the interval semantics in~\cite{ahmad2026rrt} evaluates the observed portion of the temporal window using the standard AGM rule, and obtains the lower and upper bounds by replacing each unobserved value with its worst-case and best-case completion, namely $-1$ and $1$, respectively. When the child subformula has no temporal dependence of its own, this aggregation can be updated incrementally using the modification rules in~\cite{ahmad2026rrt}, which yields an efficient monitor for non-nested STL formulae. Our method keeps this interval-monitoring framework and extends it to arbitrary nested temporal operators.

   \section{Problem Formulation}
   In this work, we study the problem of zero-shot, model-free STL planning. Specifically, we assume access to an offline dataset
   \begin{equation}
      \mathcal{D}={(\bm{x}_{t}, \bm{a}_{t}, \bm{x}_{t+1})},
   \end{equation}
   which is collected by a task-agnostic behavior policy. This dataset is used solely for representation learning, and no further interaction with the environment or task-specific retraining is permitted.

   Given an initial state $\bm{x}_{0}$ and an STL specification $\phi$ defined over the signal $\bm{s}$, the objective is to synthesize a control strategy such that the resulting trajectory satisfies $\phi$. In our framework, this objective is achieved by generating a high-level waypoint plan, which is subsequently executed by a learned goal-conditioned policy.

   \section{Overall Framework}

\begin{figure}[t]
      \centering
      \includegraphics[width=0.95\linewidth]{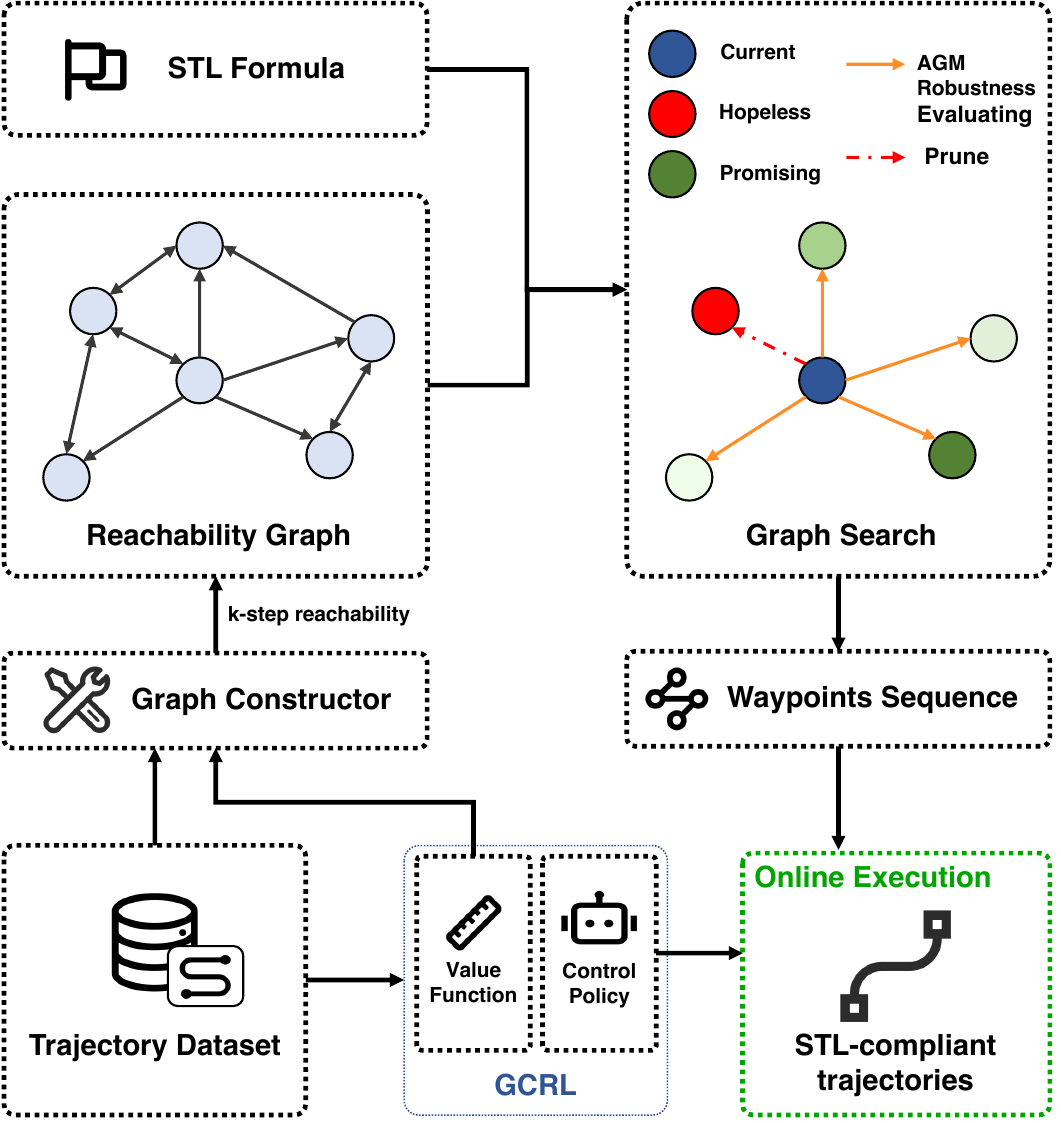}
      \caption{Overall framework of proposed GraSP-STL.
      \vspace{-10pt}}
      \label{fig:framework}
   \end{figure}

   Figure~\ref{fig:framework} illustrates the overall GraSP-STL pipeline. The framework consists of three main components:
   \begin{itemize}
      \item[(i)] an \textbf{offline goal-conditioned reinforcement learning (GCRL) agent} that provides a reachability measure (represented by a value function) and a corresponding execution policy;

      \item[(ii)] a \textbf{graph construction} module that builds a directed graph whose nodes are representative states and whose edges encode feasible $k$-step transitions under the learned measure; and

      \item[(iii)] a \textbf{graph-search} module that finds an STL-satisfying graph node sequence using AGM robustness intervals to evaluate partial plans. The resulting waypoint sequence is then executed by the learned policy.
   \end{itemize}

   Following common terminology in hierarchical planning, we refer to the graph node sequence generated by graph search as \emph{waypoints}, denoted by $\bm{w}= \left(\bm{w}_{0}, \bm{w}_{1}, \bm{w}_{2}, \bm{w}_{3}, \ldots, \bm{w}_{N}\right)$, where $\bm{w}_{i}$ is the selected graph node at time step $i$. In the ideal case, the sampled signal induced by executing the low-level controller coincides with the planned waypoint sequence, i.e., $\bm{s}= \bm{w}$, so STL satisfaction can be evaluated directly on $\bm{w}$. In practice, execution errors and approximation errors may cause deviations, i.e., $\bm{s}\neq \bm{w}$. Nevertheless, as the goal-conditioned policy is trained to realize short-horizon transitions between waypoints, the executed signal is expected to remain sufficiently close to the planned sequence, thereby ensuring that waypoint-level STL evaluation remains informative.

   \section{Goal-Conditioned Reachability Measure and Policy Learning}
   As the first component of GraSP-STL, we learn from the offline dataset $\mathcal{D}$ a goal-conditioned value function and a goal-conditioned policy. The former induces a reachability measure for planning, while the latter tracks target states with estimated transition steps within the pre-defined finite horizon. Since the graph abstraction is built from the learned measure and later executed by this policy, these two components must operate on the same transition scale.

   \textbf{Value Function Learning.} In offline goal-conditioned reinforcement learning, the value function $V(\bm{s}, \bm{g})$ measures the expected return of reaching a goal state $\bm{g}$ from a current state $\bm{s}$. Following the value-learning framework of~\cite{hiql} and the temporal-distance interpretation adopted in~\cite{ttgs}, we directly map the learned value function to an estimate of the transition steps between two states, which is then used to define a graph construction metric. Specifically, given $\bm{s}$ and $\bm{g}$, we define
   \begin{equation}
      \hat{d}(\bm{s}, \bm{g}) = - V(\bm{s}, \bm{g}),
   \end{equation}
   where $V(\bm{s}, \bm{g})$ denotes the learned goal-conditioned value function. Under this formulation, a larger value indicates that $\bm{g}$ is more reachable from $\bm{s}$ and therefore corresponds to a shorter estimated temporal distance. This interpretation is consistent with~\cite{ttgs} for graph construction and provides a reusable finite-horizon notion of reachability, which we later exploit to build the planning graph.

   \textbf{Policy Learning.} We train a goal-conditioned policy $\pi(\bm{a}\mid \bm{s}, \bm{g})$ on the same offline dataset. Inspired by the design in~\cite{baek2025graph} and further following~\cite{hiql}, the policy is designed as a direction-aware controller that executes short-horizon transitions toward target states within at most $k$ steps.

   Given state $\bm{s}$ and goal $\bm{g}$, we compute a unit direction vector from $\bm{s}$ to $\bm{g}$ and concatenate it with $\bm{s}$ as input to an MLP that parameterizes $\pi(\bm{a}\mid \bm{s}, \bm{g})$. The goal $\bm{g}$ is sampled such that the estimated transition distance $\hat{d}(\bm{s}, \bm{g}) \approx h$, where $h \sim \mathcal{U}\{1, \dots, k\}$, encouraging the policy to model $k$-step reachable transitions.

   We adopt a weighted behavior cloning objective following~\cite{hiql} but modify the weighting scheme to support directional control and encourage effective progress toward the goal. Using the learned value function $V$ and a transition sample $(\bm{s}, \bm{a}, \bm{s}')$, we define the advantage
   \begin{equation}
      A(\bm{s}, \bm{a}, \bm{g}) = V(\bm{s}', \bm{g}) - V(\bm{s}, \bm{g}),
   \end{equation}
   and a directional similarity term
   \begin{equation}
      D(\bm{s}, \bm{a}, \bm{g}) = \left\langle \frac{\bm{g}_{\text{pos}}-\bm{s}_{\text{pos}}}{\|\bm{g}_{\text{pos}}-\bm{s}_{\text{pos}}\|}, \; \frac{\bm{s}'_{\text{pos}}-\bm{s}_{\text{pos}}}{\|\bm{s}'_{\text{pos}}-\bm{s}_{\text{pos}}\|}\right\rangle,
   \end{equation}
   where $\bm{s}_{\text{pos}}$ denotes the position component of $\bm{s}$.

   The training objective is
   \begin{equation}
      \begin{aligned}
         \mathcal{L}(\pi) = -\mathbb{E}\Big[ w(\bm{s}, \bm{a}, \bm{g}) \cdot \log \pi(\bm{a}\mid \bm{s}, \bm{g}) \Big],
      \end{aligned}
   \end{equation}
   \begin{equation}
      \begin{aligned}
         w = \exp\Big( \alpha A + \beta (D - \delta) + \gamma (\|\bm{s}'_{\text{pos}}-\bm{s}_{\text{pos}}\| - \epsilon) \Big).
      \end{aligned}
   \end{equation}

   This objective favors actions that improve value, align with the goal direction, and make effective progress, ensuring consistency between value-based planning and short-horizon execution.

   \section{Graph Construction}

   Given the learned transition measure, we construct a directed graph $\mathcal{G}= (\mathcal{V}, \mathcal{E})$, where nodes correspond to representative states and edges encode feasible transitions within a control horizon $k$.

   \textbf{Node Construction.} We first construct graph nodes via a two-stage sampling and clustering procedure. Instead of directly using the full dataset, we perform a geometric grid-based uniform subsampling over the state space to improve spatial coverage. Specifically, observations are partitioned into grid cells, and a fixed number of samples are drawn approximately uniformly across occupied cells.

   On the sampled observations, we perform temporal-distance-aware clustering as in~\cite{ttgs}. Given a set of candidate states, we compute pairwise learned distances and group states whose distances fall below a threshold proportional to the horizon $k$. Each cluster is represented by its medoid, namely, the state minimizing the total intra-cluster distance. The resulting set of representative nodes $\mathcal{V}$ provides a compact abstraction that preserves both geometric coverage and consistency under the learned distance metric.

   \textbf{Edge Construction.} For each pair of nodes $\bm{v}_{i}, \bm{v}_{j}\in \mathcal{V}$, we compute the learned transition distance $\hat{d}(\bm{v}_{i}, \bm{v}_{j})$. Self-transitions are excluded by setting $\hat{d}(\bm{v}_{i}, \bm{v}_{i})=\infty$. An edge from $\bm{v}_{i}$ to $\bm{v}_{j}$ is considered feasible if
   \begin{equation}
      \hat{d}(\bm{v}_{i}, \bm{v}_{j}) < k - \delta,
   \end{equation}
   where $\delta$ is a safety margin considering estimation errors and execution uncertainties. This defines a candidate neighbor set for each node.

   To construct a sparse yet well-structured graph, we select neighbors using a directional diversity strategy. Specifically, the 2D space around each node is partitioned into angular bins, and at most one neighbor is selected per bin. Within each bin, candidates are ranked by a \emph{distance efficiency} score
   \begin{equation}
      \frac{\|\bm{v}_{j}- \bm{v}_{i}\|_{2}}{\hat{d}(\bm{v}_{i}, \bm{v}_{j})},
   \end{equation}
   which favors transitions that achieve larger geometric progress per unit learned distance.

   If the number of selected neighbors is below a target degree, additional neighbors are iteratively added by prioritizing (i) angular novelty relative to already selected edges and (ii) distance efficiency. Furthermore, edges are made approximately bidirectional: whenever $\bm{v}_{i}\to \bm{v}_{j}$ is selected and the reverse transition also satisfies the feasibility condition, the reverse edge $\bm{v}_{j}\to \bm{v}_{i}$ is added.

   \textbf{Graph Refinement.} The resulting graph may not be fully strongly connected. To ensure global reachability for planning, we extract the largest strongly connected component (SCC) and discard all other nodes and edges. The final graph $\mathcal{G}= (\mathcal{V}, \mathcal{E})$ is defined on this largest SCC.

   \section{Graph Search for STL Planning}
   Given the constructed graph $\mathcal{G}= (\mathcal{V}, \mathcal{E})$, we formulate the \emph{high-level} STL planning problem as a graph search over an augmented state space that jointly captures system evolution and STL satisfaction. Given a start state $\bm{x}_{0}$ and an STL specification $\phi$, we search for a waypoint sequence $\bm{w}= \left(\bm{w}_{0}, \bm{w}_{1}, \bm{w}_{2}, \bm{w}_{3}, \ldots, \bm{w}_{N}\right)$ whose execution by the learned goal-conditioned policy is expected to satisfy $\phi$. We do not require $\bm{x}_{0}\in \mathcal{V}$, since the initial state need not coincide with a graph node.

   \subsection{STL Robustness Evaluation}
   The main challenge in waypoint-based STL planning is to evaluate the robustness interval of a \emph{partial} waypoint sequence during graph search. We build on the incremental AGM-interval monitor in~\cite{ahmad2026rrt}, which is exact for non-nested STL formulae, and extend it to arbitrary nesting.

   The key distinction is whether the child of a temporal operator can change retroactively as new states are appended. This does not occur for non-temporal subformulae, but it does for nested temporal ones. For instance, if $\chi=\mathbf{F}_{[0,2]}\mu$, then $[\eta]_{\chi}(t)$ depends on states at $t,t+1,t+2$; hence appending a new state may tighten a previously stored value of $[\eta]_{\chi}(t)$, so a parent operator such as $\mathbf{G}_{[a,b]}\chi$ cannot be updated by a constant-time append rule alone.

   We therefore call a subformula $\psi$ \emph{immutable} if its robustness at time $t$ is fully determined once the signal at time $t$ is known. Predicates are immutable, and immutability is preserved by negation, conjunction, and disjunction. For the current partial signal $\bm{s}_{t_0:t'}$, we maintain
   \begin{equation}
      \mathcal{H}[\psi,t] = [\eta]_{\bm{s}_{t_0:t'}, \psi}(t),
   \end{equation}
   and, for temporal subformulae with immutable children, an auxiliary table $\mathcal{H}^{\mathrm{raw}}[\psi,t]$ storing the AGM aggregation over the observed part of the temporal window. At each new time step, the syntax tree of $\phi$ is traversed in post-order.

   Updates are performed as follows. Predicates and boolean operators are updated pointwise from their children. For $\psi\in\{\mathbf{G}_{[a,b]}\chi,\mathbf{F}_{[a,b]}\chi\}$ with immutable child $\chi$, we reuse the incremental update in~\cite{ahmad2026rrt}: update $\mathcal{H}^{\mathrm{raw}}[\psi,t]$ with the newly observed value and recover $\mathcal{H}[\psi,t]$ by appending the unknown suffix with $[-1,1]$. Otherwise, we recompute $\mathcal{H}[\psi,t]$ from the current child intervals over the full temporal window:
   \begin{equation}
      [\eta]_{\psi}(t)= \mathrm{AGM}_{\Box/\Diamond}\Big( \{\mathcal{H}[\chi,\tau]\}_{\tau \le t'}\cup \{[-1,1]\}_{\tau > t'}\Big),
   \end{equation}
   where $\mathrm{AGM}_{\Box/\Diamond}$ denotes $\mathrm{AGM}_{\land}$ for $\mathbf{G}$ and $\mathrm{AGM}_{\lor}$ for $\mathbf{F}$.

   Thus, our monitor reduces to~\cite{ahmad2026rrt} in the non-nested case, and remains correct for arbitrary nested STL formulae by switching to full-window reaggregation whenever immutability is violated. The procedure is summarized in Algorithm~\ref{alg:agm_update}.

   \begin{algorithm}
      [t] \small
      \caption{\textsc{EvalInterval}}
      \label{alg:agm_update} \KwIn{STL formula $\phi$, latest signal prefix $\bm{s}_{t_0:t'}$, tables $\mathcal{H}, \mathcal{H}^{\mathrm{raw}}$} \KwOut{$[\eta]_{\phi}(t_{0})$, updated $\mathcal{H}, \mathcal{H}^{\mathrm{raw}}$}

      Traverse all subformulae $\psi$ of $\phi$ in post-order\; \ForEach{subformula $\psi$ and relevant time $t$}{ \uIf{$\psi$ is a predicate or boolean operator}{ update $\mathcal{H}[\psi,t]$ pointwise from its children\; } \uElseIf{$\psi=\mathbf{G}_{[a,b]}\chi$ or $\psi=\mathbf{F}_{[a,b]}\chi$}{ \uIf{$\chi$ is immutable}{ update $\mathcal{H}^{\mathrm{raw}}[\psi,t]$ incrementally as in~\cite{ahmad2026rrt}\; recover $\mathcal{H}[\psi,t]$ by appending the unknown suffix with $[-1,1]$\; } \Else{ recompute $\mathcal{H}[\psi,t]$ by aggregating the current child intervals over the full temporal window\; } } } \Return $[\eta]_{\phi}(t_{0})$, $\mathcal{H}, \mathcal{H}^{\mathrm{raw}}$\;
   \end{algorithm}

   \emph{Complexity.} Let $|\phi|$ denote the number of subformulae and $\lVert \phi \rVert$ the temporal horizon of $\phi$. If every temporal operator has an immutable child, the per-step complexity is $\mathcal{O}(|\phi|)$, matching~\cite{ahmad2026rrt}. In the general nested case, reaggregating a temporal window costs $\mathcal{O}(\lVert \phi \rVert)$, yielding a worst-case per-step complexity of $\mathcal{O}(|\phi|\lVert \phi \rVert)$.

   \subsection{Graph Search Procedure}
   With the AGM robustness interval evaluation in place, planning reduces to a forward search over an augmented graph whose state combines a graph node with the current STL monitor state.

   \textbf{Search Node.} Each search node is defined as
   \begin{equation}
      \bm{z}= (\bm{v}, \bm{z}_{\text{parent}}, t, [\eta], \mathcal{H}, \mathcal{H}^{\text{raw}}),
   \end{equation}
   where $\bm{v}\in \mathcal{V}$ is the current graph node, $\bm{z}_{\text{parent}}$ stores the predecessor for path reconstruction, $t$ is the current time step, $[\eta]=[\underline{\eta},\overline{\eta}]$ is the robustness interval of the full specification for the partial waypoint sequence represented by $\bm{z}$, and $\mathcal{H},\mathcal{H}^{\text{raw}}$ are the monitor tables maintained by Algorithm~\ref{alg:agm_update}. Since these quantities are stored in the node itself, a frontier node never needs to be re-evaluated before expansion.

   \textbf{Initialization.} Because the initial state $\bm{x}_{0}$ need not belong to the graph, we introduce a virtual start anchor $\bm{v}_{\text{start}}$, chosen as the nearest graph node to $\bm{x}_{0}$ in geometric distance. We initialize the monitor from the singleton prefix $(\bm{x}_{0})$ and then form the frontier by extending it to each neighbor $\bm{v}'\in\mathcal{N}(\bm{v}_{\text{start}})$. Each candidate prefix $(\bm{x}_{0},\bm{v}')$ is evaluated once by Algorithm~\ref{alg:agm_update}; only candidates with $\overline{\eta}> 0$ are inserted into the frontier.

   \textbf{Successor Expansion and Pruning.} When a node $\bm{z}= (\bm{v}, \bm{z}_{\text{parent}}, t, [\eta], \mathcal{H}, \mathcal{H}^{\text{raw}})$ is selected from the frontier, the stored interval $[\eta]$ is used immediately for pruning and acceptance:
   \begin{itemize}
      \item if $\overline{\eta}\le 0$, the node is infeasible and is discarded;

      \item if $t\ge \lVert \phi \rVert$ and $\underline{\eta}>0$, the corresponding waypoint sequence satisfies the STL specification and the search terminates successfully;

      \item if $t\ge \lVert \phi \rVert$ and $\underline{\eta}\le 0$, the node is terminal but non-satisfying, so it is not expanded further.
   \end{itemize}
   Otherwise, we generate successors by appending one more graph state to the current partial waypoint sequence. For each graph neighbor $\bm{v}'\in\mathcal{N}(\bm{v})$, we evaluate the extended prefix using Algorithm~\ref{alg:agm_update} and create a child node only if the resulting upper bound remains positive. We additionally include a \emph{wait} successor, which keeps the current graph node unchanged while advancing time by one step. This action is important for temporal operators such as $\mathbf{F}$ and $\mathbf{G}$, whose satisfaction may require the passage of time without spatial motion.

   \textbf{Search Strategy.} The search proceeds over this augmented graph with any standard frontier policy, such as breadth-first, depth-first, or a robustness-based heuristic. The complete procedure is summarized in Algorithm~\ref{alg:stl_graph_search}.

   \begin{algorithm}
      [t] \small
      \caption{\textsc{STLGraphSearch}}
      \label{alg:stl_graph_search} \KwIn{Initial state $\bm{x}_{0}$, graph $\mathcal{G}=(\mathcal{V},\mathcal{E})$, STL formula $\phi$} \KwOut{A satisfying waypoint sequence or \texttt{None}}

      Initialize $(\mathcal{H}_{0}, \mathcal{H}_{0}^{raw}) \leftarrow \textsc{InitMonitor}(\bm{x}_{0})$\; Find nearest graph node $\bm{v}_{\text{start}}$ to $\bm{x}_{0}$\; Initialize frontier $\mathcal{F}\leftarrow \emptyset$\;

      \For{each $\bm{v}' \in \mathcal{N}(\bm{v}_{\text{start}})$}{ $\bm{s}_{0:1}\leftarrow (\bm{x}_{0}, \bm{v}')$\; $[\eta]', \mathcal{H}', \mathcal{H}^{raw'}\leftarrow \textsc{EvalInterval}(\phi, \bm{s}_{0:1}, \mathcal{H}_{0}, \mathcal{H}_{0}^{raw})$\; \If{$\overline{\eta}'\ge 0$}{ create node $\bm{z}'= (\bm{v}', \texttt{None}, 1, [\eta]', \mathcal{H}', \mathcal{H}^{raw'})$\; add $\bm{z}'$ to $\mathcal{F}$\; } }

      \While{$\mathcal{F}$ is not empty}{ select node $\bm{z}= (\bm{v}, \bm{z}_{\text{parent}}, t, [\eta], \mathcal{H}, \mathcal{H}^{raw})$ from $\mathcal{F}$\; \If{$t\ge \lVert \phi \rVert$ \textbf{and} $\underline{\eta}>0$}{ \Return waypoint sequence recovered from $\bm{z}$\; } \If{$t\ge \lVert \phi \rVert$ \textbf{or} $\overline{\eta}\le 0$}{ \textbf{continue}\; } \ForEach{$\bm{u}\in \mathcal{N}(\bm{v})\cup\{\bm{v}\}$}{ $\bm{s}_{0:t+1}\leftarrow$ prefix induced by $\bm{z}$ appended with $\bm{u}$\; $[\eta]', \mathcal{H}', \mathcal{H}^{raw'}\leftarrow \textsc{EvalInterval}(\phi, \bm{s}_{0:t+1}, \mathcal{H}, \mathcal{H}^{raw})$\; \If{$\overline{\eta}' \le 0$}{ \textbf{continue}\; } create node $\bm{z}'= (\bm{u}, \bm{z}, t+1, [\eta]', \mathcal{H}', \mathcal{H}^{raw'})$\; add $\bm{z}'$ to $\mathcal{F}$\; } }

      \Return \texttt{None}
   \end{algorithm}

   \textbf{Complexity.} Let $b$ denote the maximum branching factor of the reachability graph and $\lVert \phi \rVert$ the temporal horizon of the STL specification. Because the search includes both graph transitions and the explicit wait action, each node has at most $b+1$ successors. The search therefore explores an augmented state space of depth at most $\lVert \phi \rVert$, resulting in a worst-case number of $\mathcal{O}((b+1)^{\lVert \phi \rVert})$ nodes.

   At each node expansion, up to $b+1$ successors (including the wait action) are generated, and each requires an evaluation of the robustness interval via \textsc{EvalInterval}. This evaluation has complexity $\mathcal{O}(|\phi| \cdot \lVert \phi \rVert)$ in the general case with nested temporal operators, and $\mathcal{O}(|\phi|)$ when all temporal operators have immutable children.

   Therefore, the overall time complexity of the graph search procedure is $\mathcal{O}\left((b+1)^{\lVert \phi \rVert}\cdot |\phi| \cdot \lVert \phi \rVert \right)$ in the worst case, and reduces to $\mathcal{O}\left((b+1)^{\lVert \phi \rVert}\cdot |\phi|\right)$ under the assumption of immutable temporal subformulae.

   In practice, the use of robustness interval pruning ($\overline{\eta}\le 0$) significantly reduces the effective branching factor, making the search tractable for typical STL specifications.

   \textbf{Heuristics.} To further improve search efficiency, we use heuristics for frontier prioritization and dominance-based pruning. Specifically, we define a \emph{heuristic robustness interval} $[\eta_{h}]$ by modifying the AGM interval semantics in two ways.

   For atomic predicates, we use the unnormalized predicate value $h_{\mu}(\bm{s})$ instead of scaling it to $[-1,1]$, which provides finer discrimination among unsatisfied subformulae. For temporal operators, we introduce a look-ahead evaluation before the activation window. For $\psi=\mathbf{G}_{[a,b]}\chi$ and $t<a$, we define
   \begin{equation}
      \begin{cases}
         \underline{\eta}_{h\psi}(t)=\mathrm{AGM}_{\land}\big(\widetilde{\underline{\eta}}_{h\chi}(t),-1,\ldots,-1\big), \\
         \overline{\eta}_{h\psi}(t)=\mathrm{AGM}_{\land}\big(\widetilde{\overline{\eta}}_{h\chi}(t),1,\ldots,1\big),
      \end{cases}
   \end{equation}
   where the discounted child intervals are
   \begin{equation}
      \begin{aligned}
         \widetilde{\underline{\eta}}_{h\chi}(t) & =\gamma\,\underline{\eta}_{h\chi}(t)+(1-\gamma)(-1), \\
         \widetilde{\overline{\eta}}_{h\chi}(t)  & =\gamma\,\overline{\eta}_{h\chi}(t)+(1-\gamma)(1),
      \end{aligned}
   \end{equation}
   with $\gamma=\frac{1}{a-t+1}$. The case $\psi=\mathbf{F}_{[a,b]}\chi$ is defined analogously.

   Since $[\eta_{h}]$ introduces additional heuristic information, it does not preserve the soundness and convergence guarantees of the original interval semantics. Therefore, it is used only for search guidance, while the original interval $[\eta]$ is retained for pruning and acceptance, preserving the correctness of the overall search procedure.

   Based on $[\eta_{h}]$, frontier nodes are prioritized by
   \begin{equation}
      \mathrm{score}=\lambda_{0}\underline{\eta_h}+\lambda_{1}t-\lambda_{2}l,
   \end{equation}
   where $t$ is the node's time step, $l$ is the accumulated path length from the start node, and $\lambda_{0},\lambda_{1},\lambda_{2}$ are tunable.

   We further apply dominance-based pruning: for nodes sharing the same graph node and time step, only the top-$K$ candidates are retained in a fixed-size priority queue. Specifically, $\bm{z}{1}$ is said to dominate $\bm{z}{2}$ if $\underline{\eta}{1}\ge \underline{\eta}{2}$, or if $|\underline{\eta}{1}- \underline{\eta}{2}| \le \epsilon$ and $l_{1}\le l_{2}$. This pruning strategy is aggressive and may sacrifice completeness; however, it significantly reduces the search space and proves effective in practice when combined with heuristic prioritization. In most cases, pruning does not lead to failure, as the heuristic already guides the search toward promising nodes, and dominated nodes are unlikely to yield better solutions. The parameter $K$ can be tuned to balance search efficiency and solution completeness.

   \section{Experiments}

   \subsection{Experimental Setup}
   \textbf{Environments and Datasets.} We evaluate the proposed GraSP-STL framework on the \textsc{antmaze} environment of \textsc{OGBench}~\cite{park2025ogbench}, a benchmark designed for offline GCRL, with a customized maze environment illustrated in Fig.~\ref{fig:env}. The agent is an 8-DoF quadruped robot (29-dimensional state space) that can move in a 2D maze with obstacles. The offline dataset consists of 10000 trajectories, each of length 500 steps, collected using the \textsc{OGBench} data collection script with a pre-trained online goal-conditioned policy. The dataset is used for both learning the transition measure and training the low-level control policy.

   \textbf{STL Specifications.} We evaluate the framework on a diverse set of STL specifications generated from 12 templates (see Table~\ref{tab:stl_templates}), designed to cover a broad spectrum of temporal reasoning challenges, including reachability, sequencing, branching, persistence, safety constraints, and recurrent visitation objectives.

   The time bounds $t_{1}, t_{2}, t_{3}$ are chosen according to the environment scale and the agent dynamics such that all tasks are feasible yet non-trivial. For each template, we randomly sample non-overlapping circular regions in the free space with centers $(x_{i},y_{i})$ and radii $r_{i}$, and assign them to the atomic predicates $\mu_{i}$. The predicate evaluation function is defined as
   \begin{equation}
      h_{\mu_{i}}(\bm{s}) = r_{i}^{2}- \lVert \bm{s}_{\text{pos}}- (x_{i},y_{i}) \rVert^{2},
   \end{equation}
   with normalized version
   \begin{equation}
      \widetilde{h}_{\mu_{i}}(\bm{s}) = \frac{r_{i}^{2}- \lVert \bm{s}_{\text{pos}}- (x_{i},y_{i}) \rVert^{2}}{r_{i}^{2}+ \lVert \bm{s}_{\text{pos}}- (x_{i},y_{i}) \rVert^{2}}.
   \end{equation}

   \begin{table}[t]
      \caption{STL formula templates grouped by difficulty}
      \label{tab:stl_templates}
      \centering
      \renewcommand{\arraystretch}{1.3}
      \resizebox{\linewidth}{!}{
      \begin{tabular}{|c|l|}
         \hline
         \textbf{ID}                                                                      & \textbf{STL Formula}                                                                                                                                                                                                                                  \\
         \hline
         \multicolumn{2}{|c|}{\textbf{Basic} \textit{(Reachability \& Disjunction)}}       \\
         \hline
         T1                                                                               & $(\mathbf{F}_{[0, t_1]}\mu_{1}) \land (\mathbf{F}_{[t_1, t_2]}\mu_{2})$                                                                                                                                                                               \\
         T2                                                                               & $(\mathbf{F}_{[0, t_1]}\mu_{1}) \lor (\mathbf{F}_{[0, t_1]}\mu_{2})$                                                                                                                                                                                  \\
         T3                                                                               & $(\mathbf{F}_{[0, t_1]}\mu_{1}) \land (\mathbf{F}_{[0, t_1]}\mu_{2}) \land (\mathbf{F}_{[0, t_1]}\mu_{3})$                                                                                                                                            \\
         \hline
         \multicolumn{2}{|c|}{\textbf{Intermediate} \textit{(Sequential \& Safety)}}       \\
         \hline
         T4                                                                               & $(\mathbf{F}_{[0, t_1]}\mu_{1}) \land (\mathbf{F}_{[t_1, t_2]}\mu_{2}) \land (\mathbf{F}_{[t_2, t_3]}\mu_{3})$                                                                                                                                        \\
         T5                                                                               & $(\mathbf{F}_{[0, t_1]}\mu_{1}) \land (\mathbf{F}_{[t_1, t_2]}\mu_{2}) \land (\mathbf{F}_{[t_2, t_3]}\mu_{3}) \land (\mathbf{G}_{[0, t_3]}(\neg \mu_{4}))$                                                                                            \\
         T6                                                                               & $(\mathbf{F}_{[0, t_1]}\mu_{1}) \land (\mathbf{F}_{[t_1, t_2]}\mu_{2}) \land (\mathbf{F}_{[t_2, t_3]}\mu_{3}) \land (\mathbf{F}_{[t_3, t_4]}\mu_{4})$                                                                                                 \\
         T7                                                                               & $(\mathbf{F}_{[0, t_1]}\mu_{1}) \land (\mathbf{F}_{[0, t_1]}\mu_{2}) \land (\mathbf{F}_{[0, t_1]}\mu_{3}) \land (\mathbf{F}_{[0, t_1]}\mu_{4})$                                                                                                       \\
         \hline
         \multicolumn{2}{|c|}{\textbf{Advanced} \textit{(Branching, Persistence, Nested)}} \\
         \hline
         T8                                                                               & $(\mathbf{F}_{[0, t_1]}\mu_{1}) \land (\mathbf{G}_{[t_1, t_3]}\mu_{1})$                                                                                                                                                                               \\
         T9                                                                               & $(\mathbf{G}_{[t_1, t_2]}\mu_{1}) \land (\mathbf{G}_{[t_3, t_4]}\mu_{2})$                                                                                                                                                                             \\
         T10                                                                              & $\big[(\mathbf{F}_{[0, t_1]}\mu_{1}) \land (\mathbf{F}_{[t_1, t_2]}\mu_{2})\big] \lor (\mathbf{F}_{[0, t_2]}\mu_{3})$                                                                                                                                 \\
         T11                                                                              & $\big[(\mathbf{F}_{[0, t_1]}\mu_{1}) \land (\mathbf{F}_{[0, t_1]}\mu_{2})\big] \lor \big[(\mathbf{F}_{[0, t_1]}\mu_{1}) \land (\mathbf{F}_{[0, t_1]}\mu_{3})\big] \lor \big[(\mathbf{F}_{[0, t_1]}\mu_{2}) \land (\mathbf{F}_{[0, t_1]}\mu_{3})\big]$ \\
         T12                                                                              & $\mathbf{G}_{[0, t_1]}\big((\mathbf{F}_{[0, t_2]}\mu_{1}) \land (\mathbf{F}_{[t_2, t_3]}\mu_{2})\big)$                                                                                                                                                \\
         \hline
      \end{tabular}
      }
   \end{table}

   The proposed templates systematically increase in complexity. Basic tasks focus on single-step reachability and logical disjunction. Intermediate tasks introduce temporal ordering and safety constraints, requiring the agent to coordinate multiple objectives over finite horizons. Advanced templates capture more challenging temporal structures, including persistence (reach-and-stay) and branching structures (multiple feasible strategies). Importantly, we also include the most difficult class of tasks involving nested or multiple global operators (e.g., $\mathbf{G}\mathbf{F}$ specifications), which require long-horizon reasoning and repeated satisfaction of goals. For convenience, we assign an index to each template (T1--T12), which will be used to report per-template success rates and performance.

   \subsection{Case Study}
\begin{figure}[t]
      \centering
      \begin{subfigure}
         [b]{0.39\linewidth}
         \centering
         \includegraphics[width=\linewidth]{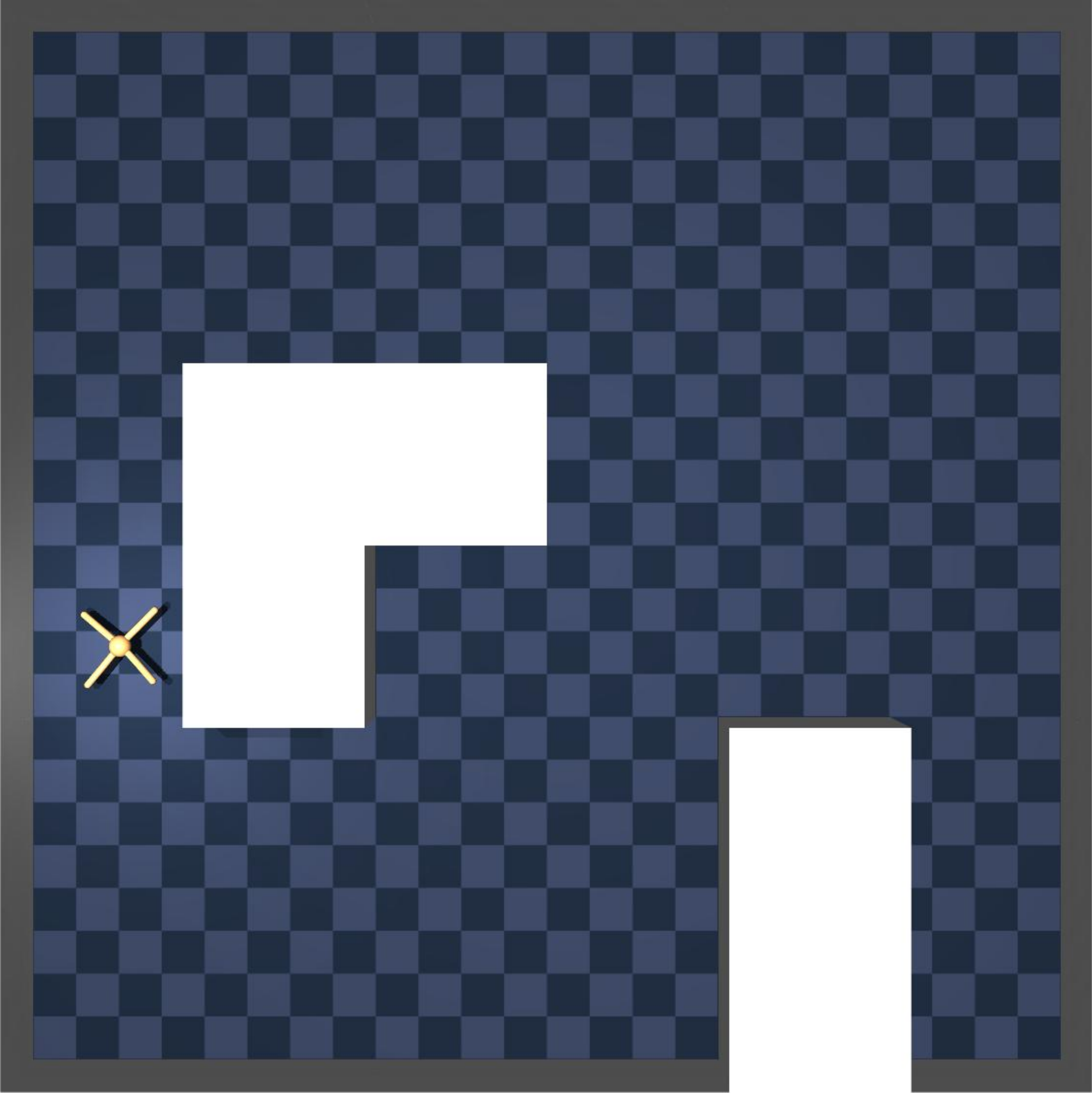}
         \caption{Environment}
         \label{fig:env}
      \end{subfigure}
      \begin{subfigure}
         [b]{0.4\linewidth}
         \centering
         \includegraphics[width=\linewidth]{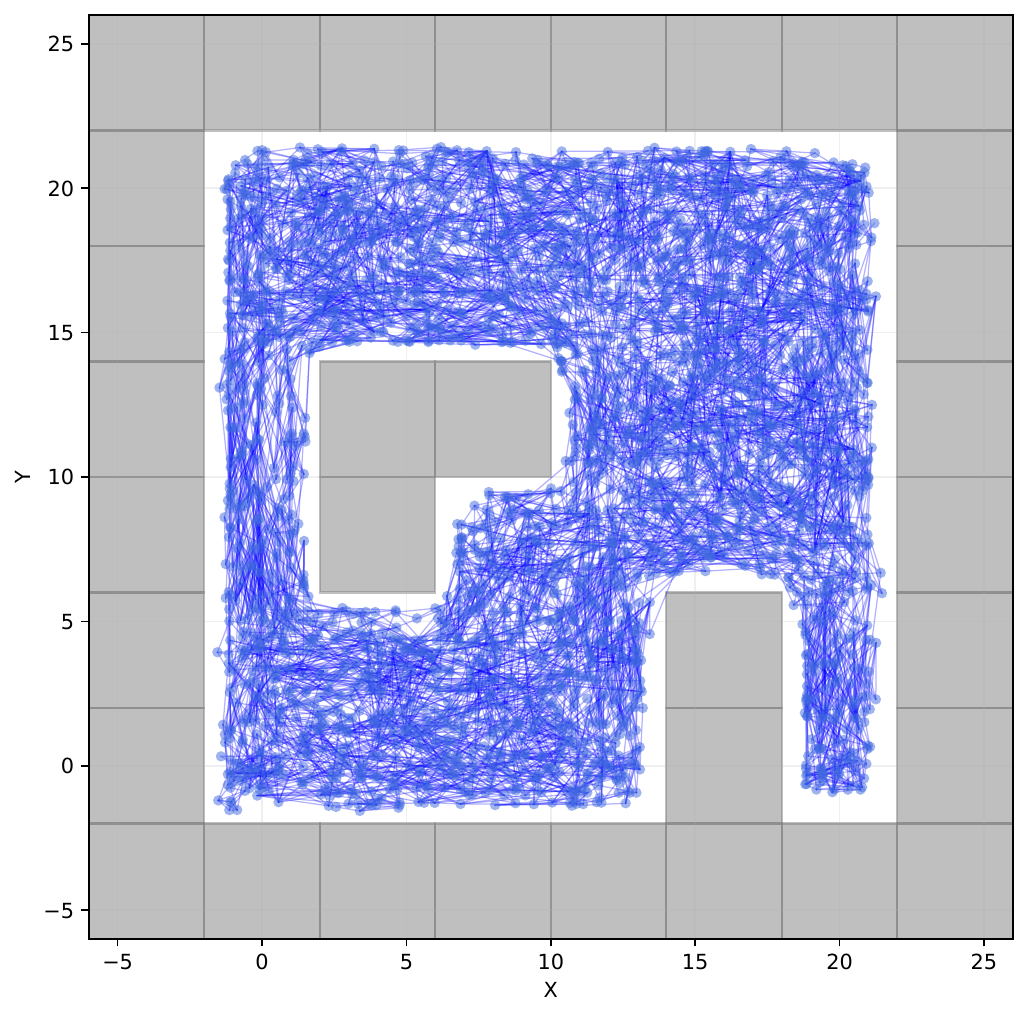}
         \caption{Graph}
         \label{fig:graph}
      \end{subfigure}

      \caption{Experimental environment and reachability graph.}
   \end{figure}

   We first present a representative case study to illustrate how GraSP-STL solves a long-horizon STL task. In our experiment, we begin by constructing a reachability graph from the dataset, which contains 3243 nodes and 8130 edges (see Fig.~\ref{fig:graph}), with an average node degree of 5.01 and an average edge length of 15.88. These properties indicate that the resulting graph achieves a favorable balance between expressiveness and computational efficiency.

   Planning is then performed on this graph under the STL specification $\phi = \mathbf{F}_{[0,12]}\mu_{1}\land \mathbf{F}_{[8,25]}\mu_{2}\land \mathbf{G}_{[20,30]}\mu_{3}$, which requires the agent to (i) reach region $\mu_{1}$ early, (ii) subsequently reach $\mu_{2}$ within an overlapping time window, and (iii) remain inside region $\mu_{3}$ during the final phase.

   \textbf{Task complexity.} This specification introduces several non-trivial challenges. The time windows of $\mu_{1}$ and $\mu_{2}$ partially overlap, allowing multiple feasible visitation orders and requiring the planner to reason over flexible temporal schedules. In addition, the persistence constraint $\mathbf{G}_{[20,30]}\mu_{3}$ enforces a terminal commitment: the agent must not only reach $\mu_{3}$ but remain there for an extended duration, creating a long-horizon dependency between earlier decisions and future feasibility. In particular, visiting $\mu_{2}$ too late may conflict with the need to settle in $\mu_{3}$ before $t=20$.

\begin{figure}[t]
      \centering
      \begin{subfigure}
         [b]{0.54\linewidth}
         \centering
         \includegraphics[width=\linewidth]{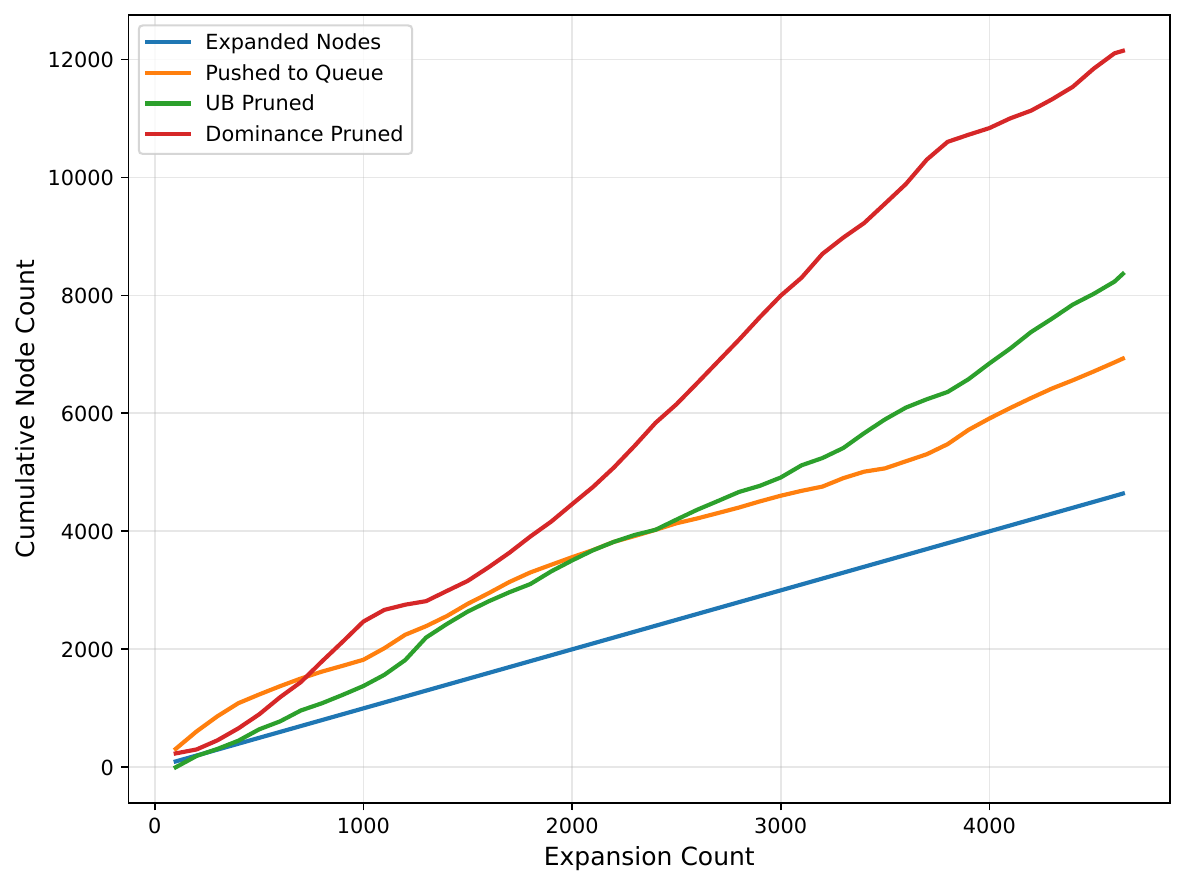}
         \caption{Planning statistics}
         \label{fig:plan_procedure}
      \end{subfigure}
      \hfill
      \begin{subfigure}
         [b]{0.44\linewidth}
         \centering
         \includegraphics[width=\linewidth]{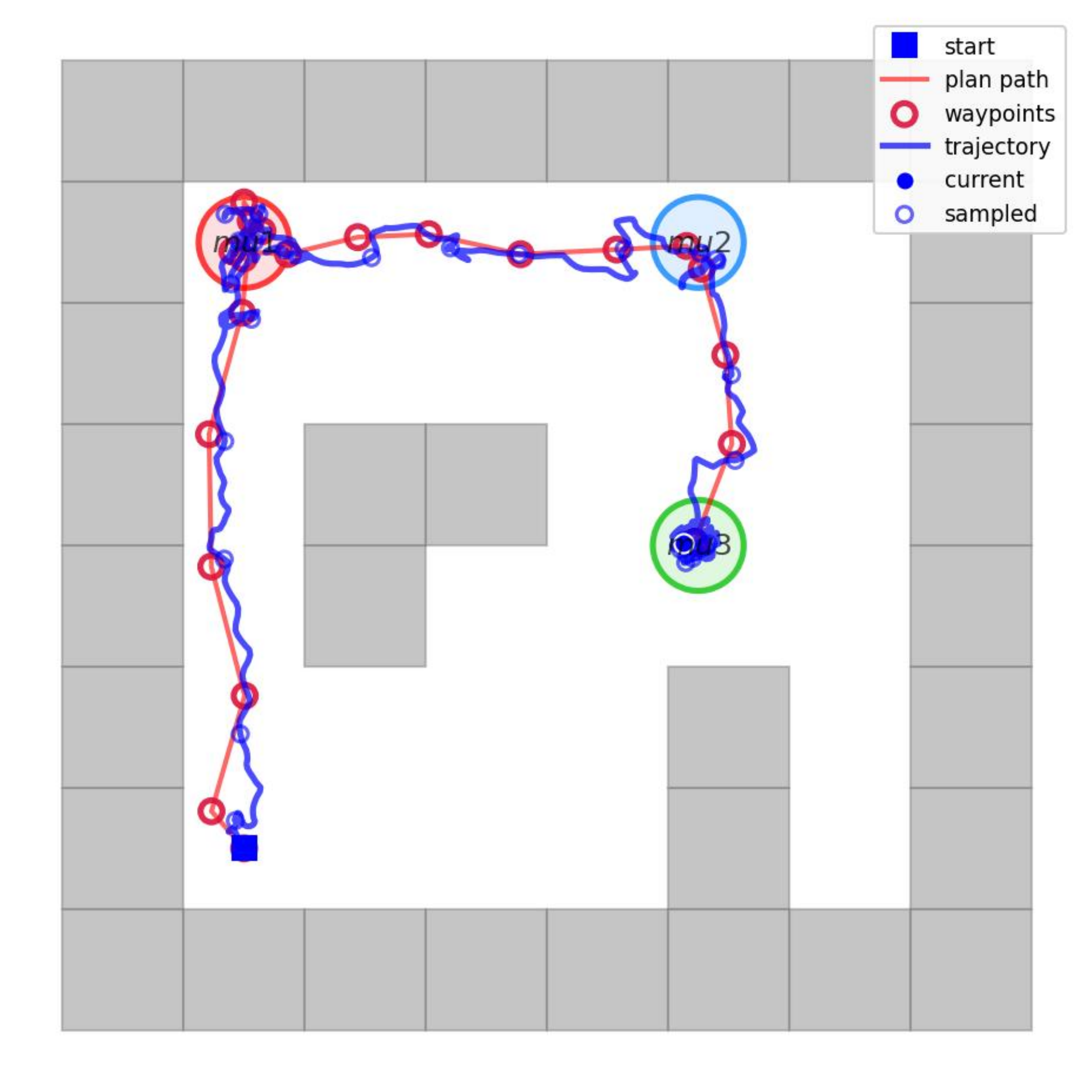}
         \caption{Execution trajectory}
         \label{fig:exec}
      \end{subfigure}

      \vspace{2ex}
      \begin{subfigure}
         [b]{0.45\linewidth}
         \centering
         \includegraphics[width=\linewidth]{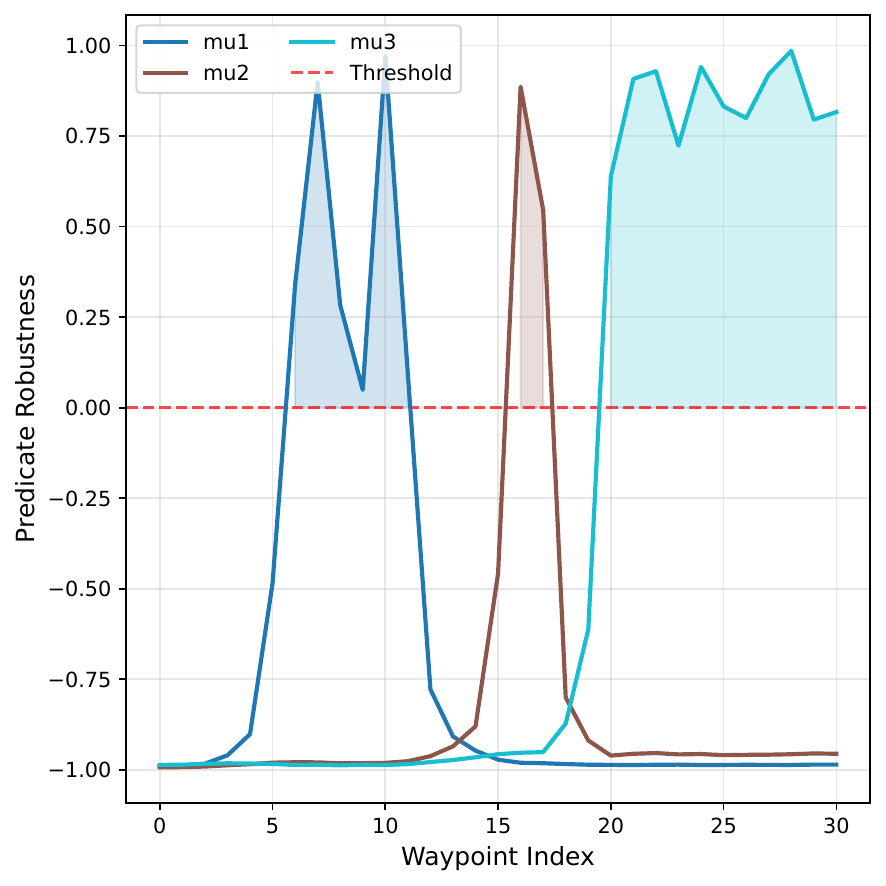}
         \caption{Robustness over predicates}
         \label{fig:pred}
      \end{subfigure}
      \hfill
      \begin{subfigure}
         [b]{0.45\linewidth}
         \centering
         \includegraphics[width=\linewidth]{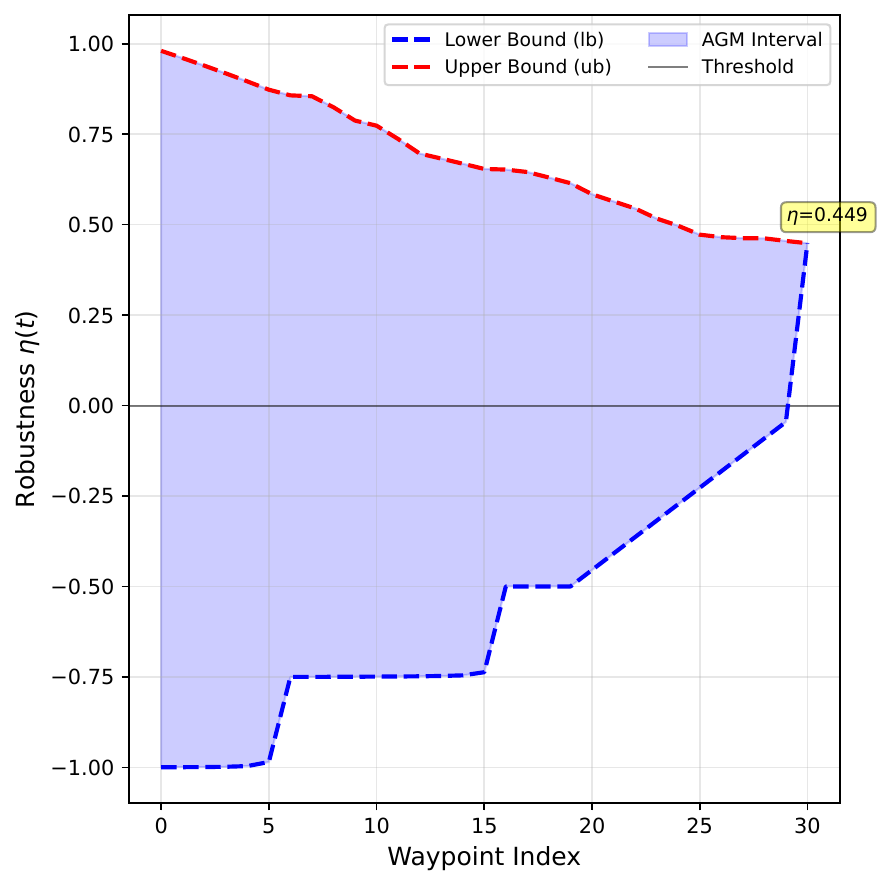}
         \caption{Robustness interval}
         \label{fig:interval}
      \end{subfigure}
      \caption{Case study of planning and execution.}
   \end{figure}

   \textbf{Planning process.} Figure~\ref{fig:plan_procedure} shows the evolution of planning statistics. As the number of expanded nodes grows, both upper-bound pruning and dominance pruning increase steadily. Notably, dominance pruning grows faster and becomes the primary pruning mechanism, indicating that many candidates are discarded due to suboptimality rather than infeasibility. Although a large number of nodes are explored, a significant portion is pruned.

   \textbf{Trajectory.} Fig.~\ref{fig:exec} shows both the planned waypoint sequence and the executed trajectory. The plan exhibits a clear temporal structure aligned with the STL specification: the agent first reaches $\mu_{1}$, then proceeds to $\mu_{2}$, and finally moves into $\mu_{3}$ and remains there. The executed trajectory closely follows the planned waypoints, and the $k$-step sampled trajectories are also well aligned, indicating that the low-level controller effectively tracks the high-level plan.

   \textbf{Robustness analysis.} As shown in Fig.~\ref{fig:pred}, $\mu_{1}$ is satisfied early, $\mu_{2}$ is achieved in the middle phase, and $\mu_{3}$ becomes strongly positive after $t \approx 20$, indicating successful entry and sustained satisfaction of the final constraint. Consistently, Fig.~\ref{fig:interval} shows that the robustness lower bound increases over time and becomes positive near the end, while the interval progressively narrows as uncertainty diminishes. This is consistent with the expected behavior of AGM interval semantics, where early stages exhibit wider intervals due to unknown future states, and later stages converge to a precise robustness value as the trajectory unfolds.

   \subsection{Large-Scale Evaluation}
   For the large-scale evaluation, we perform the proposed graph-based STL planning procedure on 2400 tasks generated from the above 12 templates, with 200 different configurations per template obtained by varying the predicate locations, predicate sizes, and starting states. This experiment is designed to assess both the scalability of the search procedure and the robustness of the learned reachability abstraction across diverse unseen STL tasks.

   \textbf{Performance Metrics.} For each template and overall, we report: \textbf{PSR(\%)}: planning success rate, \textbf{ESR(\%)}: execution success rate, \textbf{PT(s)}: planning time in seconds.

   \textbf{PSR} and \textbf{ESR} are computed over all tasks no matter the success of planning or execution, while \textbf{PT} is averaged only over successfully planned tasks. These metrics together provide a comprehensive evaluation of both the planning and execution performance under complex STL specifications. The results are summarized in Table~\ref{tab:performance_metrics}.

   \textbf{Results and Analysis.} The results highlight several strengths of the proposed framework:
   \begin{itemize}
      \item \emph{High success rates:} It achieves a planning success rate (\textbf{PSR}) of $95.58\%$ and an execution success rate (\textbf{ESR}) of $81.58\%$, and remains effective even for long-horizon tasks (up to 140 waypoints and trajectory lengths exceeding 100 units), demonstrating robustness to temporal complexity.

      \item \emph{Scalability:} although planning time increases with STL complexity, it remains tractable, with average planning time within $10$s even for the most complex cases.

      \item \emph{Waypoint accuracy:} the average distance to waypoint goals is around $0.5$, indicating good alignment with system dynamics and contributing to strong execution performance.

      \item \emph{Adaptability to branching:} for disjunctive STL specifications (e.g., T10 vs. T1/T2, T11 vs. T1), similar performance is maintained, showing effective handling of branching structures.
   \end{itemize}

   \textbf{Limitations.} Despite strong overall performance, different failure modes emerge across task types. For sequential tasks with tight temporal constraints (e.g., T6), planning success remains high while execution degrades, suggesting that temporally tight schedules leave limited slack and increase sensitivity to tracking errors. In contrast, for persistence tasks with consecutive global constraints (e.g., T9), both planning and execution success rates are lower. This is because $\mathbf{G}$ operators require the agent to enter a region at the start of the time window and remain until its end, significantly compressing the available transition time between regions. As a result, both planning feasibility and execution tolerance are reduced.

   \begin{table}[t]
      \centering
      \caption{Performance Metrics for Large-Scale Evaluation}
      \label{tab:performance_metrics}
      \renewcommand{\arraystretch}{1.15}
      \begin{tabular}{l c c c}
         \hline
         \textbf{Template}                                                                                                                                                                                                           & \textbf{PSR (\%)} & \textbf{ESR (\%)} & \textbf{PT (s)}       \\
         \hline
         \multicolumn{4}{l}{\textit{Basic (Reachability \& Disjunction)}}                                                                                                                                                             \\
         T1                                                                                                                                                                                                                          & 99.50             & 89.50             & $3.07 \pm 4.34$       \\
         T2                                                                                                                                                                                                                          & 100.00            & 97.00             & $0.80 \pm 1.35$       \\
         T3                                                                                                                                                                                                                          & 100.00            & 88.00             & $3.95 \pm 6.14$       \\
         \textbf{Basic Subtotal}                                                                                                                                                                                                     & $\bm{99.83}$      & $\bm{91.50}$      & $\bm{2.60 \pm 4.60}$  \\
         \hline
         \multicolumn{4}{l}{\textit{Intermediate (Sequential \& Safety)}}                                                                                                                                                             \\
         T4                                                                                                                                                                                                                          & 95.50             & 79.00             & $5.71 \pm 6.30$       \\
         T5                                                                                                                                                                                                                          & 94.00             & 76.00             & $10.03 \pm 11.77$     \\
         T6                                                                                                                                                                                                                          & 94.00             & 70.50             & $11.61 \pm 11.18$     \\
         T7                                                                                                                                                                                                                          & 100.00            & 79.00             & $8.41 \pm 12.74$      \\
         \textbf{Intermediate Subtotal}                                                                                                                                                                                              & $\bm{95.88}$      & $\bm{76.13}$      & $\bm{8.92 \pm 11.00}$ \\
         \hline
         \multicolumn{4}{l}{\textit{Advanced (Branching, Persistence, Combinatorial)}}                                                                                                                                                \\
         T8                                                                                                                                                                                                                          & 100.00            & 85.50             & $0.54 \pm 0.50$       \\
         T9                                                                                                                                                                                                                          & 67.50             & 54.00             & $3.48 \pm 5.18$       \\
         T10                                                                                                                                                                                                                         & 100.00            & 96.00             & $1.55 \pm 1.97$       \\
         T11                                                                                                                                                                                                                         & 100.00            & 93.50             & $4.21 \pm 5.03$       \\
         T12                                                                                                                                                                                                                         & 96.50             & 71.00             & $25.10 \pm 24.20$     \\
         \textbf{Advanced Subtotal}                                                                                                                                                                                                  & $\bm{92.80}$      & $\bm{80.00}$      & $\bm{7.09 \pm 14.78}$ \\
         \hline
         \textbf{Overall}                                                                                                                                                                                                            & $\bm{95.58}$      & $\bm{81.58}$      & $\bm{6.53 \pm 11.85}$ \\
         \hline
         \multicolumn{4}{p{0.9\linewidth}}{\footnotesize \textit{Note:} Results are presented as mean$\pm$standard deviation. \textbf{PSR}: Planning Success Rate; \textbf{ESR}: Execution Success Rate; \textbf{PT}: Planning Time.} \\
      \end{tabular}
   \end{table}

   \section{Conclusion}
   We proposed GraSP-STL, a graph-based framework for zero-shot STL planning without explicit system dynamics or task-specific retraining. By learning a transition measure from offline goal-conditioned reinforcement learning, the framework constructs a graph abstraction and performs STL planning via graph search with incremental robustness evaluation. Experimental results demonstrated strong performance on diverse long-horizon STL tasks with good computational efficiency. Despite these promising results, the current framework still depends on offline data coverage, relies on a discrete graph abstraction, and may produce temporally tight plans with limited execution slack. Future work will investigate slack-aware planning, adaptive graph refinement, and extensions to safety-critical and uncertainty-aware execution.


   \bibliographystyle{IEEEtran}
   \bibliography{reference}

@article{saxena2023funnel,
  title={Funnel-based reward shaping for signal temporal logic tasks in reinforcement learning},
  author={Saxena, Naman and Gorantla, Sandeep and Jagtap, Pushpak},
  journal={IEEE Robotics and Automation Letters},
  volume={9},
  number={2},
  pages={1373--1379},
  year={2023}
}

@inproceedings{venkataraman2020tractable,
  title={Tractable reinforcement learning of signal temporal logic objectives},
  author={Venkataraman, Harish and Aksaray, Derya and Seiler, Peter},
  booktitle={Learning for Dynamics and Control},
  pages={308--317},
  year={2020}
}

@inproceedings{wang2025multi,
  title={Multi-agent reinforcement learning guided by signal temporal logic specifications},
  author={Wang, Jiangwei and Yang, Shuo and An, Ziyan and Han, Songyang and Zhang, Zhili and Mangharam, Rahul and Ma, Meiyi and Miao, Fei},
  booktitle={IEEE/RSJ International Conference on Intelligent Robots and Systems},
  pages={6048--6054},
  year={2025}
}

@article{kurtz2022mixed, 
  title={Mixed-integer programming for signal temporal logic with fewer binary variables},
  author={Kurtz, Vincent and Lin, Hai},
  journal={IEEE Control Systems Letters},
  volume={6},
  pages={2635--2640},
  year={2022}
}

@article{yin2024formal,
  title={Formal synthesis of controllers for safety-critical autonomous systems: Developments and challenges},
  author={Yin, Xiang and Gao, Bingzhao and Yu, Xiao},
  journal={Annual Reviews in Control},
  volume={57},
  pages={100940},
  year={2024},
  publisher={Elsevier}
}

@article{belta2019formal,
  title={Formal methods for control synthesis: An optimization perspective},
  author={Belta, Calin and Sadraddini, Sadra},
  journal={Annual Review of Control, Robotics, and Autonomous Systems},
  volume={2},
  number={1},
  pages={115--140},
  year={2019},
  publisher={Annual Reviews}
}

@inproceedings{raman2014model,
  title={Model predictive control with signal temporal logic specifications},
  author={Raman, Vasumathi and Donz{\'e}, Alexandre and Maasoumy, Mehdi and Murray, Richard M and Sangiovanni-Vincentelli, Alberto and Seshia, Sanjit A},
  booktitle={IEEE Conference on Decision and Control},
  pages={81--87},
  year={2014}
}

@article{sun2022multi,
  title={Multi-agent motion planning from signal temporal logic specifications},
  author={Sun, Dawei and Chen, Jingkai and Mitra, Sayan and Fan, Chuchu},
  journal={IEEE Robotics and Automation Letters},
  volume={7},
  number={2},
  pages={3451--3458},
  year={2022}
}

@article{gu2025robust,
  title={Robust-locomotion-by-logic: Perturbation-resilient bipedal locomotion via signal temporal logic guided model predictive control},
  author={Gu, Zhaoyuan and Zhao, Yuntian and Chen, Yipu and Guo, Rongming and Leestma, Jennifer K and Sawicki, Gregory S and Zhao, Ye},
  journal={IEEE Transactions on Robotics},
  year={2025}
}

@inproceedings{vasile2017sampling,
  title={Sampling-based synthesis of maximally-satisfying controllers for temporal logic specifications},
  author={Vasile, Cristian-Ioan and Raman, Vasumathi and Karaman, Sertac},
  booktitle={2017 IEEE/RSJ International Conference on Intelligent Robots and Systems (IROS)},
  pages={3840--3847},
  year={2017}
}

@article{ahmad2026rrt,
  title={{RRT}$^\eta$: Sampling-based Motion Planning and Control from {STL} Specifications using Arithmetic-Geometric Mean Robustness},
  author={Ahmad, Ahmad and Liu, Shuo and Tron, Roberto and Belta, Calin},
  journal={arXiv:2602.16825},
  year={2026}
}

@inproceedings{mehdipour2019arithmetic,
  title={Arithmetic-geometric mean robustness for control from signal temporal logic specifications},
  author={Mehdipour, Noushin and Vasile, Cristian-Ioan and Belta, Calin},
  booktitle={American Control Conference (ACC)},
  pages={1690--1695},
  year={2019}
}

@article{hashimoto2022stl2vec,
  title={Stl2vec: Signal temporal logic embeddings for control synthesis with recurrent neural networks},
  author={Hashimoto, Wataru and Hashimoto, Kazumune and Takai, Shigemasa},
  journal={IEEE Robotics and Automation Letters},
  volume={7},
  number={2},
  pages={5246--5253},
  year={2022}
}

@article{meng2023signal,
  title={Signal temporal logic neural predictive control},
  author={Meng, Yue and Fan, Chuchu},
  journal={IEEE Robotics and Automation Letters},
  volume={8},
  number={11},
  pages={7719--7726},
  year={2023}
}

@inproceedings{stl,
  author       = {Maler, Oded and Nickovic, Dejan},
  booktitle    = {International symposium on formal techniques in real-time and fault-tolerant systems},
  organization = {Springer},
  pages        = {152--166},
  title        = {Monitoring temporal properties of continuous signals},
  year         = {2004}
}

@inproceedings{robustness,
  title={Robust satisfaction of temporal logic over real-valued signals},
  author={Donz{\'e}, Alexandre and Maler, Oded},
  booktitle={International Conference on Formal Modeling and Analysis of Timed Systems},
  pages={92--106},
  year={2010} 
}

@article{ttgs,
  title={Test-time graph search for goal-conditioned reinforcement learning},
  author={Opryshko, Evgenii and Quan, Junwei and Voelcker, Claas and Du, Yilun and Gilitschenski, Igor},
  journal={arXiv preprint arXiv:2510.07257},
  year={2025}
}

@article{hiql,
  title={Hiql: Offline goal-conditioned rl with latent states as actions},
  author={Park, Seohong and Ghosh, Dibya and Eysenbach, Benjamin and Levine, Sergey},
  journal={Advances in Neural Information Processing Systems (NeurIPS)},
  volume={36},
  pages={34866--34891},
  year={2023}
}

@inproceedings{liu2025zero,
  title={Zero-Shot Trajectory Planning for Signal Temporal Logic Tasks},
  author={Liu, Ruijia and Hou, Ancheng and Yu, Xiao and Yin, Xiang},
  booktitle={Advances in Neural Information Processing Systems 37 (NeurIPS)},
  year={2025}
}

@inproceedings{baek2025graph,
  title={Graph-Assisted Stitching for Offline Hierarchical Reinforcement Learning},
  author={Baek, Seungho and Park, Taegeon and Park, Jongchan and Oh, Seungjun and Kim, Yusung},
  booktitle={International Conference on Machine Learning (ICML)},
  pages={2391--2408},
  year={2025} 
}

@inproceedings{park2025ogbench,
  title={{OGB}ench: Benchmarking Offline Goal-Conditioned {RL}},
  author={Park, Seohong and Frans, Kevin and Eysenbach, Benjamin and Levine, Sergey},
  booktitle={International Conference on Learning Representations (ICLR)},
  pages={57515--57560},
  year={2025}
}
\end{document}